\documentclass{article}

\usepackage{microtype}
\usepackage{graphicx}
\usepackage{pgf}
\usepackage{import}
\usepackage{subcaption}
\usepackage{dblfloatfix}
\usepackage{float}
\usepackage{booktabs}

\usepackage{hyperref}
\usepackage{xurl}

\usepackage{enumitem}

\usepackage[preprint]{icml2026}

\usepackage{amsmath}
\usepackage{amssymb}
\usepackage{bm}

\icmltitlerunning{StocBench: A Benchmark for Generative Modeling of Stochastic Dynamics}

\begin{document}

\twocolumn[
  \icmltitle{StocBench: A Benchmark for Generative Modeling of Stochastic Dynamics}

  \begin{icmlauthorlist}
    \icmlauthor{Sebastian Pfister}{sch}
    \icmlauthor{Benjamin Holzschuh}{sch}
    \icmlauthor{Nils Thuerey}{sch}
  \end{icmlauthorlist}

  \icmlaffiliation{sch}{School of Computation, Information and Technology, Technical University of Munich, Germany}
  \icmlcorrespondingauthor{Sebastian Pfister}{sebastian.pfister@tum.de}
  \icmlkeywords{Diffusion models, Stochastic dynamics}

  \vskip 0.3in
]

\printAffiliationsAndNotice{}

\begin{abstract}
  We benchmark transport-based generative models as well as distillation-based few-step methods for the probabilistic forecasting of stochastic fluid flows, with a particular focus on performance under limited inference budgets.
  All methods are evaluated on a two-dimensional Kolmogorov flow with stochastic forcing. We measure one-step distributional accuracy against large simulated reference ensembles and assess whether the invariant measure is preserved during autoregressive rollouts via the enstrophy spectrum.
  On the stochastic task, flow matching achieves the most accurate one-step conditional distribution at high inference budgets, while the second-order exponential integrator \textsc{DPM-2} is strongest at very low NFE. Few-step distillation methods are competitive with the multi-step methods and preserve the enstrophy spectrum particularly well.
  A deterministic control task, in which the forcing over the prediction interval is observed, separates aleatoric from epistemic uncertainty. Model performance does not translate between the two settings: the distilled models are competitive on the stochastic task but least accurate on the control task. While stochastic diffusion samplers such as \textsc{DDPM} better preserve the enstrophy spectrum during rollouts in the stochastic setting, deterministic samplers such as \textsc{DDIM} and \textsc{DPM-2} show better spectral preservation in the deterministic setting.
  The benchmark is provided as a ready-to-use Python package at \url{https://github.com/tum-pbs/stocbench}.
\end{abstract}

\section{Introduction}
Deterministic dynamics can appear stochastic when only part of the system state is observed or when the system is highly chaotic. In these cases, we are often interested in the conditional distribution of future states rather than a single best prediction. Examples include probabilistic ensemble weather prediction~\citep{lam2023learning, price2025gencast}, stochastic subgrid-scale parameterization in climate models~\citep{berner2017stochastic}, and surrogate modeling of turbulent flows with unresolved fine-scale forcing~\citep{beck2019deep}. Benchmarks for probabilistic surrogates for stochastic PDEs (SPDEs) have recently begun to emerge: SPDEBench~\citep{li2025spdebench} evaluates neural operators across a range of SPDEs, and the concurrent TRIE framework~\citep{srikishan2026trie} proposes evaluation criteria for SPDE surrogates. Neither work, however, carefully compares transport-based generative methods, such as diffusion models, flow matching, or stochastic interpolation, or their distilled single-step variants, nor analyzes how inference budget affects distributional accuracy. \textsc{StocBench} focuses precisely on this comparison.

With this work, our aim is to provide a benchmarking scenario from the regime of turbulent flows that is comparatively easy to use, and importantly, comes with detailed ground-truth statistics and methods for evaluating distributional accuracy of different methods.

A central concept for evaluating generative models of stochastic dynamics is the \emph{invariant measure}. For a dissipative stochastic system such as stochastically forced Navier--Stokes, trajectories converge after an initial transient to a statistically stationary regime described by a probability distribution $\mu$ over the state space that is left unchanged by the dynamics. If $\omega_t \sim \mu$ and the system evolves one step forward, then likewise $\omega_{t+\Delta t} \sim \mu$. This long-time distribution encodes physically meaningful statistics, such as the average energy content at each spatial scale. A surrogate that correctly learns the single-step conditional $P(\omega_{t+\Delta t}\mid\omega_t)$ should, when composed autoregressively, produce trajectories whose long-time statistics also match $\mu$. In practice, even small biases in the learned conditional, for example systematic underestimation of variance or smoothing of high-frequency content, accumulate over rollout steps and cause the model to drift away from the true invariant measure. We evaluate this in the form of the enstrophy spectrum, which describes how vorticity energy is distributed across spatial scales at stationarity and is directly computable from ensemble rollouts.

Overall, our benchmark assesses both one-step conditional probabilistic forecasting, i.e., learning the conditional distribution of the next state given the current state, and how well the invariant measure is respected in long-horizon rollout scenarios, which is probed via the enstrophy spectrum. Specifically, we consider (i) one-step mean and standard deviation errors, as well as energy distance, and (ii) the mean difference in the average enstrophy spectrum between the predictions and ensemble references in multi-step autoregressive rollouts. All reference ensembles are obtained from a numerical simulator.

In addition to the probabilistic setting with the stochastic forcing, we evaluate a \emph{deterministic} variant in which the model is conditioned on the exact forcing.
The two variants of our benchmark probe fundamentally different model capabilities. In the \emph{stochastic} variant, the random forcing is unobserved and the uncertainty in the next state is irreducible, i.e., no additional information could remove it. The model must capture a genuine conditional distribution whose spread reflects the unknown driving noise, and autoregressive rollouts must converge to the invariant measure $\mu$. In the \emph{deterministic} variant, the forcing is provided as input and the next state is fully determined by the current vorticity and forcing together. The task therefore reduces to learning a deterministic simulation step, closely resembling the regime studied by neural operators~\citep{li2020fourier,brandstetter2022message,holzschuh2025pde}. A well-calibrated model should collapse its predictive distribution to a point mass in this setting. The two tasks are thus complementary: the stochastic task tests distributional generation and long-time spectral statistics, while the deterministic task acts as a sanity check that models suppress spurious stochasticity when the physics is fully observed. Additionally, comparing the ranking of models in the stochastic and deterministic settings allows us to answer the question: \textbf{Does model performance translate from the stochastic setting to the deterministic one and vice versa?}

For both variants, we evaluate several transport-based models on our benchmark, including DDPM \citep{ho2020denoising}, DDIM \citep{song2020denoising}, DPM-2 \citep{cheng2022dpm}, flow matching \citep{lipman2023flow}, and stochastic interpolation \citep{albergo2022building,albergo2025stochastic,chen2024probabilistic}. Since transport-based models often perform poorly when the inference budget is very small, we also evaluate distillation methods that are designed to generate samples in only a few steps, or even a single step \citep{song2023consistency,sauer2024adversarial}.
Throughout, we focus on \emph{single-step} conditional methods that learn $P(\omega_{t+\Delta t}\mid\omega_t)$. Restricting to this setting isolates the quality of the generative step itself from confounding factors such as temporal bundling or extended input histories; we discuss these extensions in the related work.

Importantly, StocBench provides a statistically robust evaluation. The training set comprises 1,500 trajectories with 200 frames each, totaling 300,000 turbulent-flow snapshots. For the one-step evaluation, we estimate ground-truth conditional statistics from 5,000 simulated next states for each of 48 conditions, and compare them against model ensembles of 3,000 samples per condition. For the long-term evaluation, we unroll 5,000 initial states over 50 autoregressive steps, yielding 250,000 predicted transitions from which we assess whether models preserve the enstrophy spectrum of the invariant measure.

To summarize, our contributions are:
\begin{itemize}[leftmargin=*,topsep=2pt,itemsep=1pt]
  \item We introduce \textsc{StocBench}, a benchmark for conditional stochastic forecasting in fluid dynamics based on the stochastically forced Kolmogorov flow of \citet{chen2024probabilistic}. The benchmark provides datasets and evaluation metrics in a ready-to-use \href{https://github.com/tum-pbs/stocbench}{Python package} for reproducible model comparison. Special care was taken to provide robust ground truth statistics for evaluation.
  \item We evaluate representative transport-based generative methods on \textsc{StocBench} and analyze their performance under inference budgets ranging from 10 to 400 network evaluations per sample. We further compare them against distillation-based approaches that generate samples with only one or two network evaluations.
  \item We include a deterministic prediction task in which the model is conditioned on the forcing, yielding dynamics and enstrophy spectra similar to the stochastic variant. This task separates aleatoric from epistemic uncertainty and tests whether model performance translates between the stochastic and deterministic settings.
\end{itemize}

\section{Related Work}

\paragraph{Neural PDE and SPDE benchmarks.} 
Benchmark studies for neural PDE solvers have historically focused on deterministic dynamics and architecture comparisons, using short-horizon prediction error as the primary metric~\citep{li2020fourier,lu2019deeponet}. PDE-Refiner~\citep{lippe2023pde} extends rollout stability of neural PDE solvers by applying iterative denoising refinements, but remains in the deterministic regime. For stochastic dynamics, SPDEBench~\citep{li2025spdebench} provides a large-scale evaluation comparing architectures including FNO~\citep{li2020fourier}, Neural SPDEs~\citep{salvi2022neural}, and DLR-Net across noise regimes and numerical discretizations, but emphasizes architectural differences over distributional accuracy. For turbulent flows, \citet{kohl2026benchmarking} benchmark autoregressive conditional diffusion models on incompressible flows, assessing accuracy and diversity of generated trajectories. \citet{harder2025efficient} study probabilistic surrogate modeling for partially observed dynamical systems. Most directly related is the concurrent TRIE framework~\citep{srikishan2026trie}, which proposes three evaluation criteria for stochastic PDE surrogates: reproduction of invariant measures, calibrated uncertainty estimates via CRPS, and computational efficiency, and compares stochastic interpolants against deterministic neural architectures on stochastic Kolmogorov flow and the Kuramoto--Sivashinsky equation. \textsc{StocBench} shares the focus on distributional evaluation but distinguishes itself by systematically comparing the full class of transport-based generative methods and distillation approaches under controlled inference budgets.

\paragraph{Transport-based generative models for physical dynamics.} 
Score-based diffusion models~\citep{ho2020denoising, song2020score} and flow matching~\citep{lipman2023flow,liu2022flow} have seen broad application in physical simulation and forecasting. For weather prediction, \citet{lam2023learning} demonstrate that deterministic neural models can match numerical weather prediction, while probabilistic variants generate calibrated ensemble forecasts~\citep{price2025gencast}. Stochastic interpolation~\citep{albergo2022building,chen2024probabilistic,albergo2025stochastic} provides a unified framework for constructing stochastic transport processes between arbitrary distributions, including between consecutive simulation snapshots. All these methods share a common structure. They map a simple prior to the target distribution via an ODE or SDE, and converge to the correct distribution as the number of integration steps increases. \textsc{StocBench} provides a controlled evaluation of their relative strengths under limited inference budgets.

\paragraph{Few-step and single-step inference.} 
Consistency models~\citep{song2023consistency,song2023improved} and adversarial diffusion distillation~\citep{sauer2024adversarial} compress multi-step transport into one or a few network evaluations by distilling a multi-step teacher. While distillation methods trade theoretical convergence guarantees for inference speed, this trade-off is important for ensemble-based probabilistic forecasting, which requires many independent samples. \citet{chen2025convergence} show theoretically that the benefit of additional consistency sampling steps depends on the rate at which training error decreases, explaining the empirically observed saturation of quality gains beyond two steps. We include both distillation approaches in our benchmark to assess whether they can match or exceed transport-based methods under the limited inference budgets where the latter are most disadvantaged.

\paragraph{Multi-frame prediction and extended temporal context.} 
The $t_i \to t_{i+1}$ formulation we study is the most fundamental and tightly controlled setting for benchmarking generative models of stochastic dynamics, but two natural extensions are often employed in practice. First, models can be trained to predict multiple future frames $\{t_{i+1}, \ldots, t_{i+n}\}$ jointly in a single forward pass, which is known as temporal bundling \citep{brandstetter2022message}. Temporal bundling addresses \emph{exposure bias}, i.e., a model trained on single-step predictions is applied at inference time to its own outputs rather than ground-truth inputs, and small prediction errors compound over long rollouts. Predicting $n$ steps jointly aligns the training and inference distributions more closely, reducing this compounding effect. Second, models can condition on an extended input history $\{t_{i-k}, \ldots, t_i\}$ rather than only the current state, providing additional temporal context that can reduce prediction uncertainty and improve long-term physical consistency~\citep{shysheya2024conditional}. Diffusion Forcing~\citep{chen2024diffusionforcing} unifies these ideas in a probabilistic framework. By applying independent per-token noise levels across an entire multi-step sequence, it enables joint generation over full trajectories with diffusion-style iterative refinement and flexible conditioning on any subset of past frames. \textsc{StocBench} deliberately restricts to the $t_i \to t_{i+1}$ setting to provide the most controlled comparison of generative modeling and distillation strategies, isolating the quality of individual generative steps from the confounding effects of temporal bundling or extended history conditioning. Extending the benchmark to cover these richer formulations is a natural direction for future work.

\section{Benchmark}
Our benchmark evaluates models on two variants of Kolmogorov flow. Stochastic Kolmogorov flow assesses whether generative models can reproduce the stochasticity induced by unobserved random forcing, while deterministic Kolmogorov flow tests whether the models make deterministic predictions when the forcing is observed. Dataset generation details for both variants are provided in Appendix~\ref{app:dataset}.

\begin{figure}[t]
  \centering
  \resizebox{\linewidth}{!}{\import{plots/dataset_preview/}{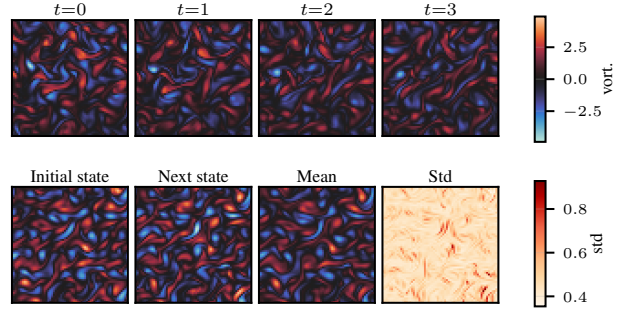}}
  \caption{Dataset preview for stochastic Kolmogorov flow. First row: consecutive vorticity snapshots. Second row: next-state ensemble statistics for a fixed initial condition, with the mean and standard deviation estimated from 3000 samples.}
  \label{fig:dataset_preview}
\end{figure}

\subsection{Stochastic Kolmogorov Flow}
\label{sec:stoc_kf}
Following \citet{chen2024probabilistic}, we simulate the incompressible Navier--Stokes equations on the two-dimensional torus $\mathbb{T}^2=[0,2\pi]^2$ with stochastic forcing. For $\bm{x}=(x,y)\in\mathbb{T}^2$ and $t\in[0,T]$, the vorticity field $\omega(\bm{x},t)$ evolves according to the vorticity--streamfunction formulation
\begin{equation}
  d\omega = \left[ -(\bm{v}\cdot\nabla)\omega + \nu\Delta\omega -\alpha\omega \right] dt + \epsilon d\eta,
  \label{eq:ns_sde}
\end{equation}
where $\bm{v}$ is the velocity field, $\nu>0$ is the kinematic viscosity, $\alpha>0$ is a linear drag (Rayleigh friction) coefficient that dissipates large-scale energy, and $\epsilon>0$ scales the stochastic forcing. The forcing increment $d\eta$ is a linear combination of fixed basis functions scaled by the differentials of independent Wiener processes $W_{\bm{k}}$ and $V_{\bm{k}}$:
\begin{equation}
  d\eta(\bm{x},t) = \sum_{\bm{k}\in\mathcal{K}}\sin(\bm{k}\cdot\bm{x})\,dW_{\bm{k}}+\cos(\bm{k}\cdot\bm{x})\,dV_{\bm{k}},
  \label{eq:forcing}
\end{equation}
where $\mathcal{K} = \{(6,0), (7,0), (5,5), (8,8)\}$ is the set of forced wavenumbers.

The forecasting task is to learn the conditional next-state distribution $P(\omega_{t+\Delta t}\vert\omega_t)$, where uncertainty arises from the unobserved forcing realization over the interval $[t,t+\Delta t]$. Figure~\ref{fig:dataset_preview} shows consecutive vorticity snapshots together with the next-state ensemble statistics for a fixed initial condition.

\subsection{Deterministic Kolmogorov Flow}
\label{sec:det_kf}
In the second variant, we make the prediction step deterministic by conditioning on the forcing. Rather than using Wiener increments, we sample a forcing $\eta_t$ for each snapshot interval $[t,t+\Delta t]$ and hold it fixed throughout the interval. Given the current vorticity $\omega_t$ and the applied forcing $\eta_t$, the next vorticity $\omega_{t+\Delta t}$ is deterministic up to numerical solver accuracy.

\begin{figure}[t]
  \centering
  \resizebox{\linewidth}{!}{\import{plots/enstr_spec_comp/}{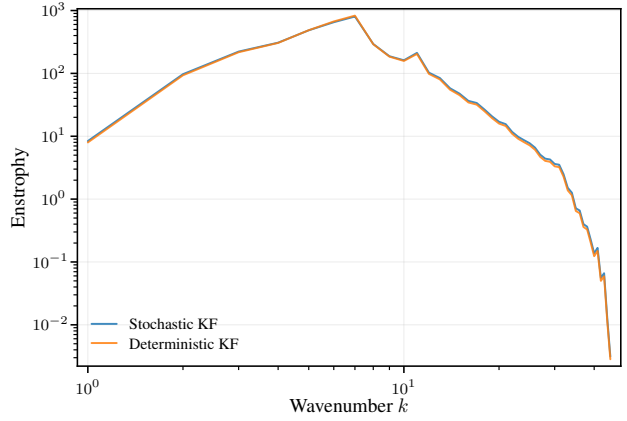}}
  \caption{Both Kolmogorov-flow variants have very similar average radial enstrophy spectra.}
  \label{fig:enstr_spec_variants}
\end{figure}

Although the stochastic variant is driven by a Wiener process, whereas the deterministic variant keeps the forcing fixed over each snapshot interval, their average enstrophy spectra are very similar, as shown in Figure~\ref{fig:enstr_spec_variants}.

\subsection{Metrics}
\label{sec:metrics}
In the stochastic setting, the current state $\omega_t$ does not uniquely determine the next state, but instead defines a conditional distribution $P(\omega_{t+\Delta t} \mid \omega_t)$. To assess how well each baseline model distribution $Q_\theta(\omega_{t+\Delta t} \mid \omega_t)$ approximates the true next-snapshot distribution $P$, we use three one-step metrics; a fourth metric evaluates whether the invariant measure is preserved during autoregressive rollouts:
\begin{enumerate}
  \item \textbf{Conditional Mean Error} ($\epsilon_{\mu}$): Quantifies the error in the expected next state. Let $\mu_P$ and $\mu_Q$ denote the pointwise ensemble means of the true distribution $P$ and the predicted distribution $Q_\theta$, respectively. The relative $L_2$ error is
    \[\epsilon_{\mu}=\frac{\|\mu_P-\mu_Q\|_2}{\|\mu_P\|_2}.\]
  \item \textbf{Conditional Standard Deviation Error} ($\epsilon_{\sigma}$): Quantifies the error in the predicted stochastic spread. Let $\sigma_P$ and $\sigma_Q$ denote the corresponding pointwise conditional standard deviations. The relative $L_2$ error is
    \[\epsilon_{\sigma}=\frac{\|\sigma_P-\sigma_Q\|_2}{\|\sigma_P\|_2}.\]
  \item \textbf{Energy Distance} ($\epsilon_D$): Quantifies the mismatch between the predicted and true next-snapshot conditional distributions beyond their first two moments. For a grid point $(i,j)$, the energy distance between the model distribution and the true distribution is defined as
      \[D_{i,j}=2\,\mathbb{E}\bigl[|x-y|\bigr]-\mathbb{E}\bigl[|x-x'|\bigr]-\mathbb{E}\bigl[|y-y'|\bigr],\]
      where $x,x'$ are independent samples from $Q_\theta$ at grid point $(i,j)$, and $y,y'$ are independent samples from $P$ at the same grid point. We define
      \[\epsilon_D=\frac{1}{N_xN_y}\sum_{i=1}^{N_x}\sum_{j=1}^{N_y}D_{i,j}\]
    as the spatial average of the pointwise energy distance, estimated from ensembles of generated and ground-truth samples. This metric is closely related to the continuous ranked probability score (CRPS; see, e.g., \citealp{gneiting2007strictly}) and is used when the full reference distribution is available rather than a single observed outcome.
  \item \textbf{Enstrophy Spectrum Error} ($\epsilon_Z$): Quantifies how well the invariant measure $\mu$ is preserved during autoregressive rollouts. Starting from an ensemble of states drawn from $\mu$, we compare the ensemble-averaged radial enstrophy spectra of the model, $\hat{Z}_t(k)$, and the solver, $Z_t(k)$, at each rollout step $t$ and radial wavenumber $k$. The mean absolute error over $T_{\mathrm{roll}}$ rollout steps and $K$ radial bins is
  \[\epsilon_Z=\frac{1}{T_{\mathrm{roll}}K}\sum_{t=1}^{T_{\mathrm{roll}}}\sum_{k=1}^{K}\left|Z_t(k)-\hat{Z}_t(k)\right|.\]
\end{enumerate}

\section{Baselines}
\label{sec:baselines}

Most multi-step generative models can be formulated as continuous-time transport processes governed by stochastic differential equations (SDEs) or ordinary differential equations (ODEs) \citep{song2020score}. This unified framework makes the baselines directly comparable. By varying the number of solver steps, we can analyze the trade-off between distributional fidelity and inference cost. We quantify inference cost by the number of function evaluations (NFEs), i.e., the number of network forward passes required to generate a sample.

Throughout this section, $t$ denotes physical time and $\tau$ the transport time of the generative process. Diffusion models transport noise at $\tau=1$ to data at $\tau=0$, whereas flow matching and stochastic interpolation run in the opposite direction, from noise at $\tau=0$ to data at $\tau=1$. All baselines learn $P(\omega_{t+\Delta t} \mid \omega_t)$ and receive the conditioning state $\omega_t$ as an additional network input, which we suppress in the notation.

\subsection{Diffusion Models}
\label{sec:diffusion}

Diffusion models learn to reverse a forward process that gradually corrupts data with Gaussian noise \citep{ho2020denoising}. In the limit of infinitely many diffusion steps, the forward process can be described by an SDE of the form
\begin{equation}
    dx = f(x, \tau)\, d\tau + g(\tau)\, dw,
    \label{eq:sde_fwd}
\end{equation}
where $\tau \in [0,1]$ is the diffusion time, $f(x,\tau)$ is the drift coefficient, $g(\tau)$ is the diffusion coefficient, and $w$ is a Wiener process \citep{song2020score}. We adopt the variance-preserving (VP) noise schedule $\beta(\tau)$ from DDPM \citep{ho2020denoising}, corresponding to $f(x,\tau) = -\tfrac{1}{2}\beta(\tau)x$ and $g(\tau) = \sqrt{\beta(\tau)}$. The reverse process is described by the SDE
\begin{equation}
    dx = \left[ f(x,\tau) - g(\tau)^2 \nabla_x \log p_\tau(x) \right] d\tau + g(\tau)\, d\bar{w},
    \label{eq:sde_rev}
\end{equation}
where $p_\tau$ denotes the marginal density of the forward process at diffusion time $\tau$ and $\bar{w}$ is a Wiener process running backward in time. The Fokker--Planck equation further implies that every such diffusion process admits a deterministic probability flow ODE with the same marginals $p_\tau(x)$ \citep{song2020score}, which enables deterministic sampling. This ODE is given by
\begin{equation}
    dx = \left[ f(x,\tau) - \tfrac{1}{2} g(\tau)^2 \nabla_x \log p_\tau(x) \right] d\tau.
    \label{eq:prob_flow_ode}
\end{equation}
In both the reverse SDE \eqref{eq:sde_rev} and the probability flow ODE \eqref{eq:prob_flow_ode}, the score function $\nabla_x \log p_\tau(x)$ is the only unknown term. In practice, it is expressed in terms of the noise predictor $\epsilon_\theta(x,\tau)$ using the approximation $\nabla_x \log p_\tau(x_\tau) \approx -\epsilon_\theta(x_\tau,\tau)/\sigma_\tau$, where $\sigma_\tau$ is the standard deviation of the forward perturbation kernel $p_\tau(x_\tau \mid x_0)$ around a clean sample $x_0$. The network $\epsilon_\theta$ is trained to predict the noise added by the forward process \citep{ho2020denoising}.

Sampling draws from the standard Gaussian prior and integrates either the SDE \eqref{eq:sde_rev} or the ODE \eqref{eq:prob_flow_ode} from $\tau=1$ to $\tau=0$. Our benchmark compares three samplers that share a single trained $\epsilon_\theta$-prediction model and differ only in the numerical integration scheme:
\begin{itemize}
    \item \textsc{DDIM} \citep{song2020denoising} is a first-order exponential integrator for the ODE \eqref{eq:prob_flow_ode}: the linear drift is integrated exactly, and $\epsilon_\theta$ is held constant at its value at the start of each step. Each step requires one NFE.
    \item \textsc{DPM-2} is the second-order exponential integrator DPM-Solver-2 \citep{cheng2022dpm}: the linear drift is again exact, and the $\epsilon_\theta$ integral is approximated by a midpoint rule. Each step therefore requires two NFEs.
    \item \textsc{DDPM} \citep{ho2020denoising} performs ancestral sampling: each update draws from the Gaussian posterior of the discretized forward process, with the clean state replaced by its $\epsilon_\theta$-based estimate. This corresponds to a first-order stochastic discretization of the reverse SDE \eqref{eq:sde_rev} \citep{song2020score}. Each step requires one NFE.
\end{itemize}

\subsection{Flow Matching}
\label{sec:fm}

Flow matching \citep{lipman2023flow} learns a time-dependent vector field $u_\tau(x)$ whose flow transports a Gaussian prior to the target distribution. Because this marginal vector field is intractable, \citet{lipman2023flow} introduce conditional flow matching: a neural network $v_\theta$ is regressed onto the conditional vector field $u_\tau(x \mid x_1)$, which is available in closed form for a fixed data sample $x_1$. \citet{lipman2023flow} show that this objective has the same minimizer as regression onto the intractable marginal field.

We use the conditional optimal transport (OT) path, which interpolates linearly between noise $x_0 \sim \mathcal{N}(0, I)$ and a data sample $x_1$, with conditional vector field
\begin{equation}
    u_\tau(x \mid x_1) = \frac{x_1 - (1-\sigma_{\min}) x}{1 - (1-\sigma_{\min})\tau},
    \label{eq:ot_path}
\end{equation}
where $\sigma_{\min} > 0$ is a small constant controlling the residual noise at $\tau=1$ (see Table~\ref{tab:hparams}). Along this path, each conditional trajectory is a straight line traversed at constant speed. In practice, this also results in straighter marginal trajectories than those of the probability flow ODE \eqref{eq:prob_flow_ode}, so fewer steps suffice for accurate integration.

To generate samples, we draw noise from the prior and numerically integrate the learned ODE defined by $v_\theta$ from $\tau=0$ to $\tau=1$. For our FM baseline, we use the standard Euler method, at a cost of one NFE per step.

\subsection{Stochastic Interpolation}
\label{sec:si}

Stochastic interpolation defines a time-dependent process $I_\tau$ that smoothly transitions between a base observation $x_0$ and a target $x_1$:
\begin{equation}
    I_\tau = \alpha_\tau x_0 + \beta_\tau x_1 + \sigma_\tau W_\tau, \qquad \tau \in [0,1].
    \label{eq:si_interpolant}
\end{equation}
Here, $W = (W_\tau)_{\tau \in [0,1]}$ is a Wiener process, and $\alpha_\tau$, $\beta_\tau$, and $\sigma_\tau$ are differentiable scalar schedules satisfying boundary conditions that ensure $I_0 = x_0$ and $I_1 = x_1$; specifically, $\alpha_0 = \beta_1 = 1$ and $\alpha_1 = \beta_0 = \sigma_1 = 0$. In contrast to the previous methods, both endpoints are physical states: $x_0 = \omega_t$ is the conditioning state and $x_1 = \omega_{t+\Delta t}$ the next snapshot, so the process bridges consecutive states instead of transporting from a Gaussian prior.

\citet{chen2024probabilistic} show that, for $x_1 \sim q(x_1 \mid x_0)$, there exists an SDE whose marginal distributions match those of the interpolation process $I_\tau$. This SDE is defined by
\begin{equation}
    dX_\tau = b_\tau(X_\tau, x_0)\, d\tau + \sigma_\tau\, dW_\tau,
    \label{eq:si_sde}
\end{equation}
where the drift coefficient $b_\tau(x, x_0)$ can be learned by minimizing the regression objective
\begin{equation}
    \mathbb{E}_{\tau, x_0, x_1, W} \left[ \left\| b_\tau(I_\tau, x_0) - R_\tau \right\|^2 \right],
    \label{eq:si_objective}
\end{equation}
with target vector field $R_\tau = \dot{\alpha}_\tau x_0 + \dot{\beta}_\tau x_1 + \dot{\sigma}_\tau W_\tau$.

For our SI baseline, we follow \citet{chen2024probabilistic} and use their interpolant schedules (Table~\ref{tab:hparams}). We then numerically integrate the corresponding SDE \eqref{eq:si_sde} using the Euler--Maruyama method, at a cost of one NFE per step.

\subsection{Consistency Distillation}
\label{sec:cd}

Consistency models \citep{song2023consistency} enforce a self-consistency constraint across different noise levels of the same denoising trajectory, enabling sampling in far fewer steps. We formalize this constraint with a consistency function $f: \mathbb{R}^d \times [0,T] \to \mathbb{R}^d$ that maps a noisy state $x_\tau$ at noise time $\tau$ into data space. For two states $x_\tau$ and $x_{\tau'}$ lying on the same denoising trajectory, consistency requires
\begin{equation}
    f(x_\tau, \tau) = f(x_{\tau'}, \tau').
    \label{eq:consistency}
\end{equation}
Together with the boundary condition $f(x_0, 0) = x_0$, this ensures that every point along a given trajectory maps to the exact same clean sample $x_0$.

Consistency distillation trains $f_\theta$ by enforcing this constraint along the probability flow ODE of a pretrained teacher diffusion model rather than learning a consistency function from scratch. This teacher-guided objective is generally more stable and yields higher sample quality. Sampling draws from the prior and evaluates $f_\theta$ at the maximum noise level. For multi-step generation, this prediction is refined by alternately adding noise and re-evaluating at progressively lower noise levels. In our benchmark, we evaluate consistency distillation using single- and multi-step sampling.
\begin{itemize}
    \item \textsc{CD-1}: Evaluates the consistency function once at the maximum noise level.
    \item \textsc{CD-2}: Uses one additional evaluation at the noise level that is most effective for the specific forecasting task (see Appendix~\ref{app:cd_schedules}).
\end{itemize}

\subsection{Adversarial Diffusion Distillation}
\label{sec:add}

Adversarial Diffusion Distillation (ADD) \citep{sauer2024adversarial} extends the standard generative adversarial network (GAN) setup \citep{goodfellow2020generative}, consisting of a generator and a discriminator, by introducing a frozen, pretrained teacher model. The discriminator learns to distinguish between real and generated samples, assigning high scores to real data and low scores to generated data. The generator optimizes a dual objective: a distillation loss that minimizes the discrepancy between student and teacher samples, and an adversarial loss that maximizes the discriminator score. Importantly, because of adversarial guidance, the generator is not strictly limited by the sampling quality of the teacher \citep{sauer2024adversarial}.

While ADD originally uses diffusion models, our ADD-FM baseline adapts the framework to flow matching using a one-step generator and a 10-step teacher. Sampling therefore requires a single NFE. Training details are given in Appendix~\ref{app:add}.

\section{Experiments}
\begin{figure*}[tb]
  \centering
  \resizebox{\linewidth}{!}{\import{plots/stoc_errors/}{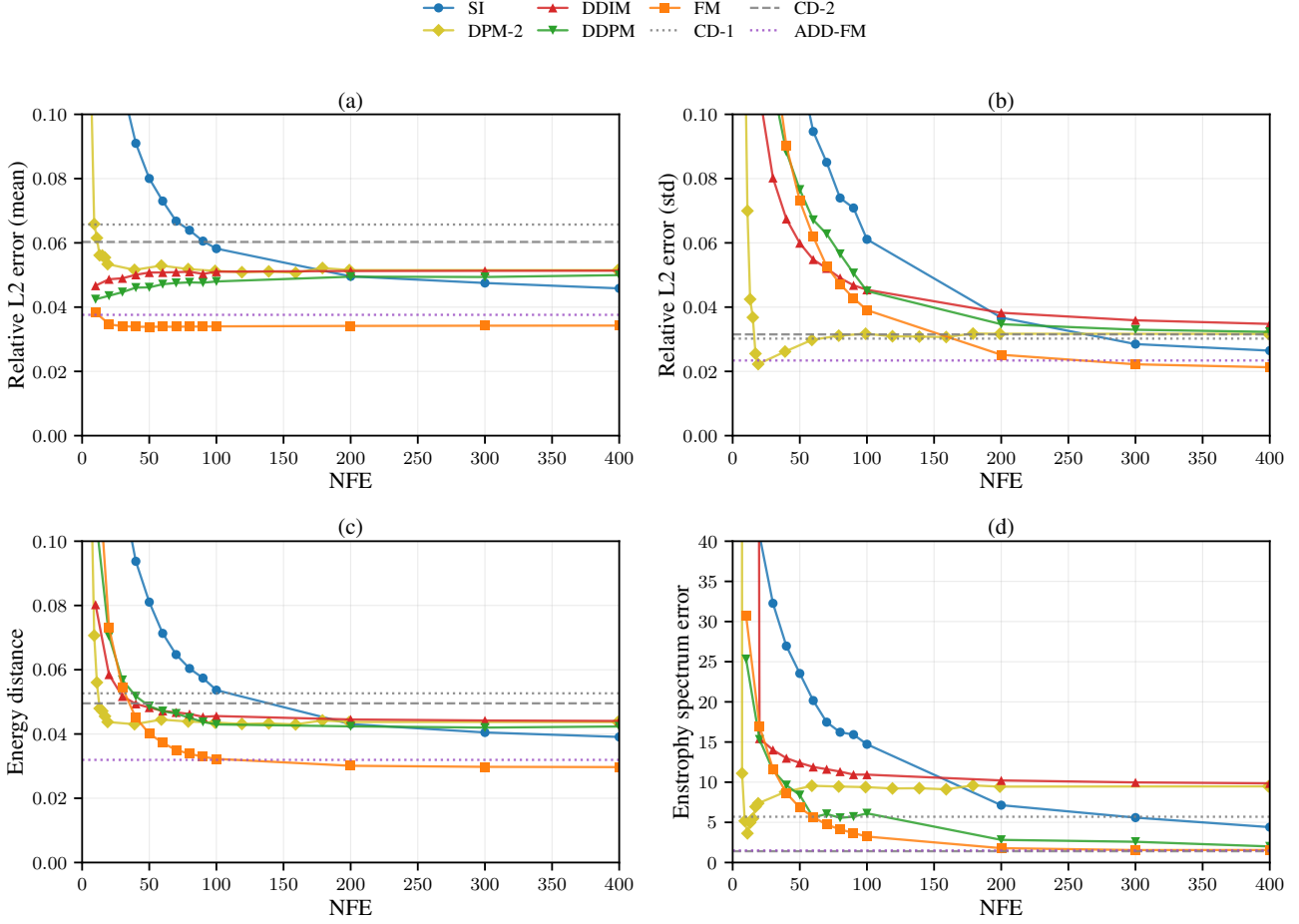}}
  \caption{Metrics for stochastic Kolmogorov flow that measure one-step distributional fit (a--c) as well as preservation of the invariant measure $\mu$ during autoregressive rollout, assessed in the enstrophy spectrum (d). All metrics are defined in Section~\ref{sec:metrics}. Lower values indicate closer agreement with the reference distribution. Markers denote the inference budget while dotted and dashed lines indicate models using 1 and 2 NFEs, respectively.}
  \label{fig:stoc_errors}
\end{figure*}

We organize the experiments into two parts. First, we evaluate all baselines on \emph{stochastic} Kolmogorov flow using the metrics defined in Section~\ref{sec:metrics}. This setting tests whether the models can learn dynamics with inherent stochasticity. Second, we evaluate the same baselines on the \emph{deterministic} variant to assess whether they produce accurate and deterministic predictions when the conditioning information fully determines the next state.

All models are trained on the full training set and evaluated on 48 initial conditions with an ensemble size of 3,000; further details are given in Appendix~\ref{app:dataset}. During evaluation, we use the exponential moving average (EMA) version of each model with a decay rate of $0.999$. Implementation details for each model are provided in Appendix~\ref{app:baselines}.

\subsection{Stochastic Kolmogorov Flow}
\label{sec:results_stoc}

\begin{figure*}[tb]
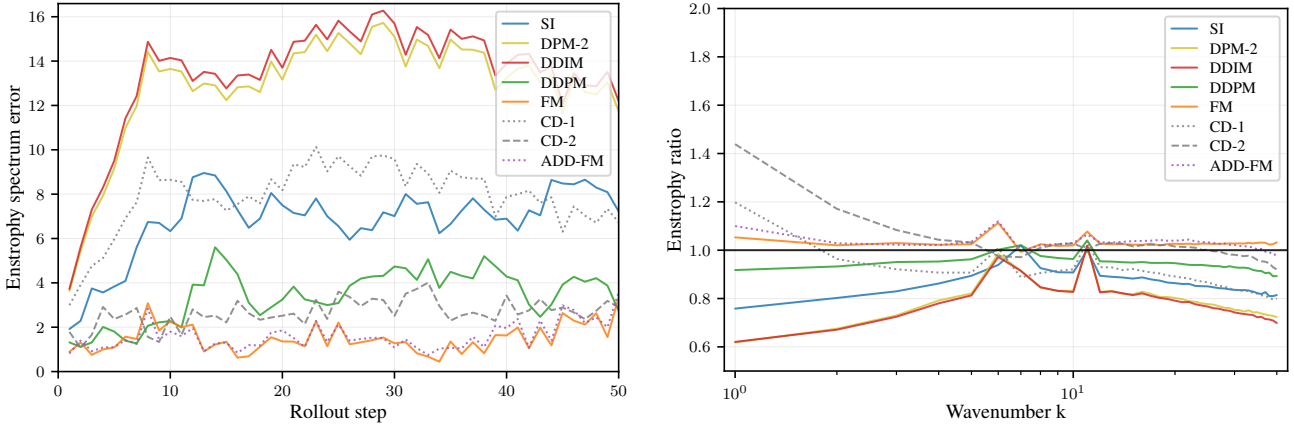

  \centering
  \begin{subfigure}[t]{0.49\textwidth}
    \centering
    \resizebox{\linewidth}{!}{\import{plots/stoc_enstr_rollout/}{stoc_enstr_rollout.pgf}}
  \end{subfigure}
  \hfill
  \begin{subfigure}[t]{0.49\textwidth}
    \centering
    \resizebox{\linewidth}{!}{\import{plots/stoc_enstr_ratio/}{stoc_enstr_ratio.pgf}}
  \end{subfigure}
  \caption{Evaluation of model invariant measures for stochastic Kolmogorov flow, obtained from autoregressive rollouts and assessed by the enstrophy spectrum. \emph{Left}: enstrophy spectrum error at each step of a 50-step rollout. \emph{Right}: ratio of predicted to reference enstrophy after 50 rollout steps; values above the black line indicate overestimation, while values below indicate underestimation. In both subplots, dotted, dashed, and solid lines denote models using 1, 2, and 400 NFEs, respectively.}
  \label{fig:stoc_enstr_rollout}
\end{figure*}

The stochastic task evaluates whether the methods can reproduce the conditional next-state distribution induced by unobserved forcing. We assess one-step distributional fit with mean and standard deviation errors, as well as energy distance, which captures differences beyond the first two moments. We also evaluate whether the models preserve the invariant measure $\mu$. To do so, we initialize autoregressive rollouts from an ensemble of states sampled from $\mu$ and compare the enstrophy spectrum after rollout with that of $\mu$.

In terms of mean error, flow matching (\textsc{FM}) performs best across all inference budgets (Figure~\ref{fig:stoc_errors}a). This behavior is consistent with flow matching learning comparatively straight transport trajectories, whereas diffusion-based methods such as \textsc{DDPM}, \textsc{DDIM}, and \textsc{DPM-2} follow more curved probability-flow paths. At higher inference budgets, stochastic interpolation (\textsc{SI}) begins to outperform the other diffusion methods, yet still leaves a gap to \textsc{FM}. Since \textsc{SI} interpolates between consecutive simulation snapshots, its endpoints are typically close in state space, resulting in less curved stochastic bridges than trajectories initialized from a generic Gaussian distribution. Among the distilled models, adversarial diffusion distillation (\textsc{ADD-FM}) achieves a substantially lower mean error than consistency distillation, likely due to its strong flow matching teacher. Adding a second NFE at an optimized noise level (\textsc{CD-2}) slightly improves the mean error over \textsc{CD-1} (Appendix~\ref{app:cd_schedules}). Figure~\ref{fig:stoc_seed_fields} shows the corresponding conditional mean and standard deviation error fields for three example conditions.

For the standard deviation error in Figure~\ref{fig:stoc_errors}b, \textsc{DPM-2} performs particularly well at low inference budgets. Its error exhibits a sharp minimum at only 20 NFEs, where \textsc{DPM-2} outperforms all other multi-step baselines. This minimum likely stems from coarse discretization effects that compensate for the model's systematic underestimation of the standard deviation (Figure~\ref{fig:std_norm}, left). For higher inference budgets, \textsc{FM} achieves the lowest standard deviation error, with \textsc{SI} slowly approaching. Notably, \textsc{ADD-FM} matches the performance of \textsc{FM} with a single inference step.

Figure~\ref{fig:std_norm} (left) shows one-step \emph{predictive variability}, calculated as the spatial $L^2$ norm of the pointwise ensemble standard deviation. Most methods underestimate variability at low inference budgets and approach the ground truth as the number of NFEs increases. The only exception is \textsc{DPM-2}, which overestimates the conditional standard deviation at low budgets.

The invariant-measure evaluations (Figures~\ref{fig:stoc_errors}d and~\ref{fig:stoc_enstr_rollout}) suggest that deterministic denoising increases spectral damping. Deterministic sampling acts as a spectral filter, whereas stochastic sampling preserves small-scale energy through the injected noise. This behavior is most evident for the diffusion models, which share the same $\epsilon$-prediction model. \textsc{DDIM} and \textsc{DPM-2} converge to invariant measures with lower enstrophy than the ground-truth reference, whereas \textsc{DDPM} retains most of the enstrophy. \textsc{FM} is the exception among the multi-step methods, achieving the lowest enstrophy spectrum error at high inference budgets, with an enstrophy error that remains nearly constant throughout the 50-step rollout (Figure~\ref{fig:stoc_enstr_rollout}, left). The corresponding spectra after 50 rollout steps are shown in Figure~\ref{fig:stoc_enstr_spec}.

\subsection{Deterministic Kolmogorov Flow}
\label{sec:results_det}

\begin{figure*}[t]
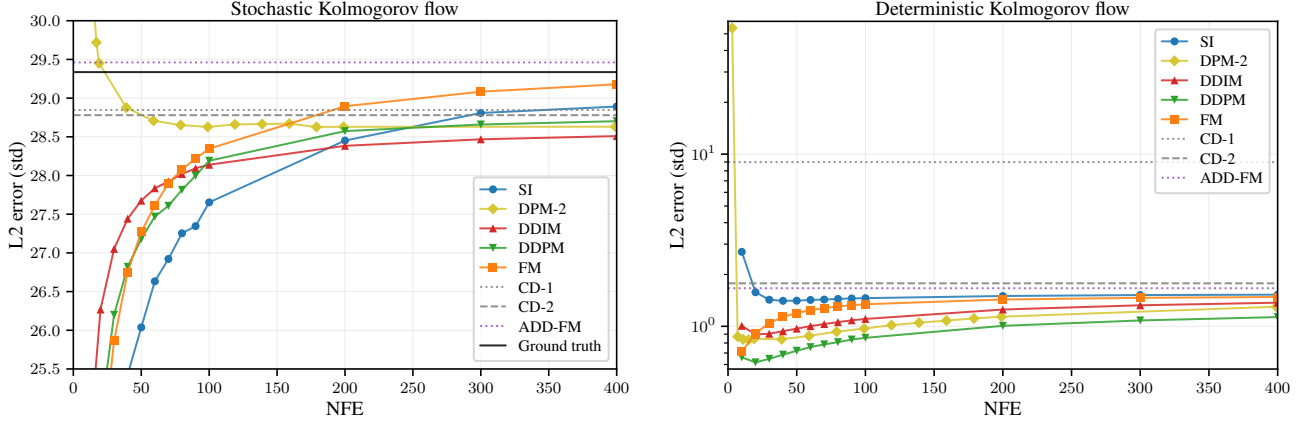

  \centering
  \begin{subfigure}[t]{0.49\textwidth}
    \centering
    \resizebox{\linewidth}{!}{\import{plots/stoc_std_norm/}{stoc_std_norm.pgf}}
  \end{subfigure}
  \hfill
  \begin{subfigure}[t]{0.49\textwidth}
    \centering
    \resizebox{\linewidth}{!}{\import{plots/det_std_norm/}{det_std_norm.pgf}}
  \end{subfigure}
  \caption{Variability of one-step predictions for the stochastic (left) and deterministic (right) Kolmogorov flow as a function of NFE. Markers denote the inference budget while dotted and dashed lines indicate models using 1 and 2 NFEs, respectively.}
  \label{fig:std_norm}
\end{figure*}
\begin{figure*}[t]
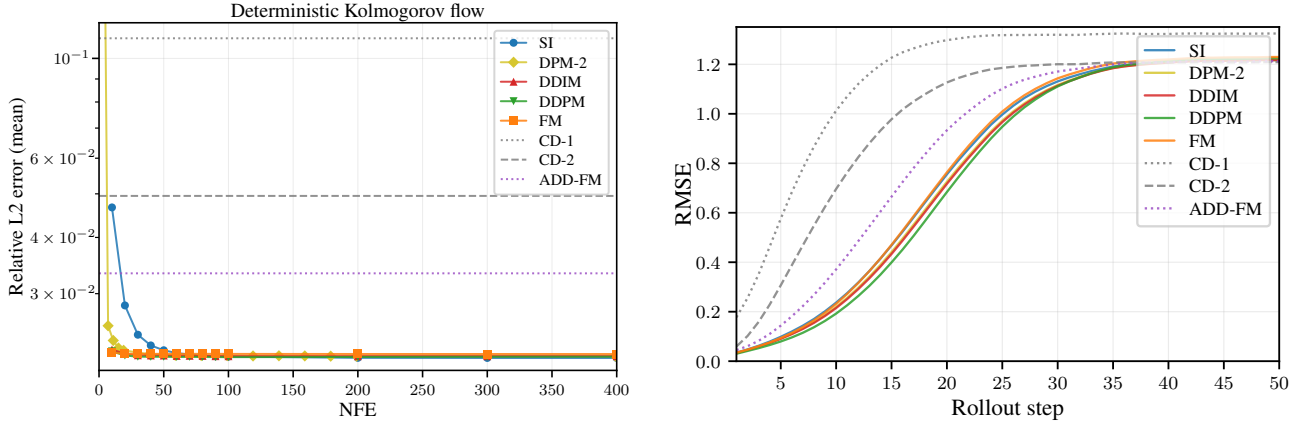

  \centering
  \begin{subfigure}[t]{0.49\textwidth}
    \centering
    \resizebox{\linewidth}{!}{\import{plots/det_mean_err/}{det_mean_err.pgf}}
  \end{subfigure}
  \hfill
  \begin{subfigure}[t]{0.49\textwidth}
    \centering
    \resizebox{\linewidth}{!}{\import{plots/det_rmse_rollout/}{det_rmse_rollout.pgf}}
  \end{subfigure}
  \caption{Accuracy on deterministic Kolmogorov flow relative to the numerical solver. \emph{Left}: For each inference budget, we report the error between the ensemble mean of the one-step predictions and the deterministic ground-truth next snapshot generated by the numerical solver. Markers denote the inference budget, with dotted and dashed lines corresponding to 1 and 2 NFEs, respectively. \emph{Right}: RMSE at each rollout step, averaged over autoregressive trajectories initialized from randomly sampled states. Markers denote the inference budget, while dotted, dashed, and solid lines correspond to evaluations using 1, 2, and 400 NFEs, respectively.}
  \label{fig:det_errors}
\end{figure*}

\begin{figure*}[t]
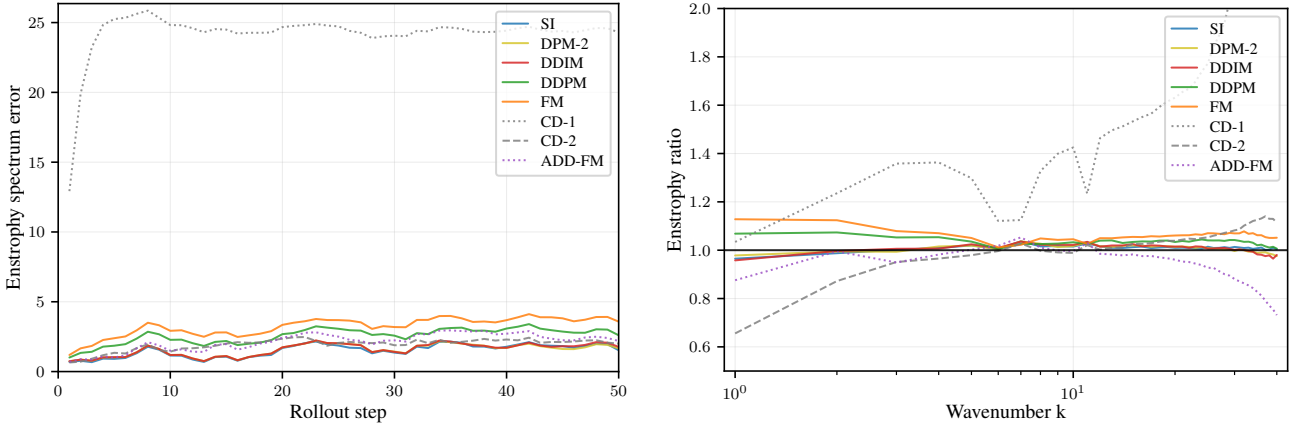

  \centering
  \begin{subfigure}[t]{0.49\textwidth}
    \centering
    \resizebox{\linewidth}{!}{\import{plots/det_enstr_rollout/}{det_enstr_rollout.pgf}}
  \end{subfigure}
  \hfill
  \begin{subfigure}[t]{0.49\textwidth}
    \centering
    \resizebox{\linewidth}{!}{\import{plots/det_enstr_ratio/}{det_enstr_ratio.pgf}}
  \end{subfigure}
  \caption{Evaluation of model invariant measures for deterministic Kolmogorov flow, obtained from autoregressive rollouts and assessed by the enstrophy spectrum. \emph{Left}: enstrophy spectrum error at each rollout step. \emph{Right}: ratio of predicted to reference enstrophy after 50 rollout steps; values above the black line indicate overestimation, while values below indicate underestimation. In both subplots, dotted, dashed, and solid lines denote models using 1, 2, and 400 NFEs, respectively.}
  \label{fig:det_enstr_rollout}
\end{figure*}

To assess whether the predictive variability observed on the stochastic task reflects the intrinsic stochasticity of the dynamics, we evaluate all models on a deterministic control task. Predictive variability has two possible sources: \emph{aleatoric} uncertainty stems from the stochastic
forcing and is irreducible, whereas \emph{epistemic} uncertainty stems from imperfect learning of the dynamics and would vanish for a perfect model. The control task eliminates the aleatoric source by conditioning the models on the forcing over the next snapshot interval, so that the next state is fully determined. Any residual variability is therefore epistemic. Figure~\ref{fig:std_norm} reports the predictive variability for both tasks.

Overall, single- and few-step distillation methods exhibit higher residual variability than multi-step methods (Figure~\ref{fig:std_norm}, right), indicating greater epistemic uncertainty in predicting the next step with only one or a few inference steps. Among the single-step methods, \textsc{CD-1} shows substantially higher residual variability than \textsc{ADD-FM}. This contrasts with the stochastic setting, where \textsc{CD-1} accurately recovers the conditional distribution using a single inference step. As shown in Appendix~\ref{app:cd_schedules}, this discrepancy arises from the sampling schedule: for stochastic dynamics, a single evaluation at the highest noise level is sufficient, whereas deterministic dynamics require additional lower-noise evaluations to eliminate residual variability.

The mean error shows an even larger gap between distillation and multi-step methods (Figure~\ref{fig:det_errors}, left). The multi-step methods form a tight cluster with low errors, while the distilled models have significantly higher errors. Figure~\ref{fig:det_errors} (right) evaluates long-term accuracy by autoregressively unrolling an ensemble of randomly sampled initial conditions for 50 steps and reporting the ensemble-averaged RMSE relative to the deterministic solver trajectory. Single- and few-step distillation methods deviate faster from the solver trajectory, while all models maintain stable rollouts over the full rollout horizon.

Figure~\ref{fig:det_enstr_rollout} shows that preserving spatial structures during long autoregressive rollouts is easier in the deterministic setting than in the stochastic setting. Except for \textsc{CD-1}, all ensemble-averaged enstrophy spectra remain close to the reference spectrum. Unlike in the stochastic setting, distillation methods do not consistently outperform multi-step methods. In particular, \textsc{CD-1} exhibits excessive enstrophy, especially at high-frequency modes (Figure~\ref{fig:det_enstr_rollout}, right). The absolute spectra after 50 rollout steps are shown in Figure~\ref{fig:det_enstr_spec}.

\section{Discussion}
The results paint an interesting picture of the pros and cons of different generative samplers as numerical instruments for stochastic physics:
\begin{itemize}
\item Flow matching (\textsc{FM}) is the most accurate method at high inference budgets on the stochastic task: it achieves the best one-step conditional distribution and, among the multi-step samplers, best preserves the enstrophy spectrum.
\item \textsc{DPM-2} has advantages in the \textit{low-budget} regime: its exponential integrator yields accurate conditional distributions with only a few NFEs, especially for the standard deviation (Figure~\ref{fig:stoc_errors}b).
\item The three diffusion samplers \textsc{DDIM}, \textsc{DPM-2} and \textsc{DDPM} share a single trained model and differ only in the sampling procedure, yet on the stochastic task their enstrophy spectrum errors over autoregressive rollouts differ by a factor of five: \textsc{DDIM} and \textsc{DPM-2} converge to a damped invariant measure, whereas \textsc{DDPM} preserves it (Figure~\ref{fig:stoc_enstr_rollout}). On the deterministic task this ranking reverses (Figure~\ref{fig:det_enstr_rollout}). The stochastic updates of \textsc{DDPM} reintroduce small-scale enstrophy, which is beneficial in the stochastic setting but results in excessive enstrophy in the deterministic one. The one-step metrics of the diffusion samplers are very similar, so one-step accuracy alone does not imply spectral accuracy over long rollouts.
\item Distillation methods can be highly competitive, with \textsc{ADD-FM} in particular showing strong preservation of the invariant measure at a single NFE (Figure~\ref{fig:stoc_enstr_rollout}). However, their training often does not converge monotonically, and part of their predictive variability is epistemic, i.e., it stems from imperfect learning rather than the stochastic forcing (Figure~\ref{fig:std_norm}, right).
\item Stochastic interpolation (\textsc{SI}) is intuitive, as it bridges nearby simulation states rather than transporting from pure noise. Nevertheless, it needs a relatively large inference budget to become competitive.
\end{itemize}
Our benchmark shows that no single method dominates all regimes. Flow matching is the most accurate method on the stochastic task, but only at high inference budgets. Distillation methods are competitive using only one or two NFEs on the stochastic task, yet fall behind all multi-step samplers in the deterministic setting. \textsc{DPM-2} is the strongest multi-step method at low budgets and performs consistently in both settings, though its accuracy plateaus early and does not improve at higher budgets.

\section{Conclusion}
We have introduced \textsc{StocBench}, a benchmark for evaluating generative modeling of stochastic dynamics. Which method is preferable depends on the inference budget: flow matching is the most accurate at high budgets, while the distilled models are competitive at a single NFE on the stochastic task. At low budgets, \textsc{DPM-2} is the strongest multi-step method and, unlike the distilled models, does not suffer from unstable training.

We find that performance does not translate between the stochastic setting and that of the deterministic control task. The distilled models are competitive on the stochastic task but least accurate on the control task, suggesting that part of their predictive variability is epistemic, i.e., caused by imperfect learning rather than the inherent stochasticity of the dynamics. Among the diffusion samplers, the ranking is reversed between the two settings: the stochastic sampler \textsc{DDPM} best preserves the rollout enstrophy spectrum on the stochastic task, whereas the deterministic samplers \textsc{DDIM} and \textsc{DPM-2} preserve it better on the deterministic one.

While we focused on making \textsc{StocBench} as accessible as possible to guide the development of future generative models and distillation algorithms for stochastic dynamics, future work could provide benchmark scenarios for higher dimensions, resolutions, and other stochastic systems.

\clearpage

\bibliography{paper_dblp,paper_doi}
\bibliographystyle{icml2026}

\clearpage
\appendix
\onecolumn
\section{Dataset Generation}
\label{app:dataset}

\subsection{Stochastic Kolmogorov Flow}
\paragraph{Training.}
The training dataset consists of 1,500 trajectories, each containing 200 frames. We simulate the vorticity by integrating Equation~\eqref{eq:ns_sde} using a pseudo-spectral solver with Euler--Maruyama time stepping. Each simulation starts from zero vorticity and runs for 100 warm-up snapshot intervals to allow the state distribution to converge toward the invariant measure $\mu$. Simulations are performed on a $256\times 256$ grid and subsequently reduced to $64\times 64$ through spatial averaging. All trajectories are normalized using the global mean and standard deviation to improve training stability. Table~\ref{tab:sim_params} summarizes the physical constants and simulation parameters.

\paragraph{Testing.}
For one-step evaluation, we generate 48 conditioning states and, for each, an ensemble of 5,000 next-snapshot states. These ensembles are used to estimate the ground-truth conditional statistics (e.g., mean and standard deviation) against which we compare the model-generated ensemble statistics. For the invariant-measure evaluation, we generate 5,000 additional states as starting points for autoregressive rollouts.

\subsection{Deterministic Kolmogorov Flow}

\paragraph{Training.}
For the deterministic variant, we use the same Kolmogorov-flow dynamics as in Equation~\eqref{eq:ns_sde}, but replace the Wiener forcing with a piecewise-constant forcing that is held fixed between consecutive snapshots. This forcing is provided to the model as conditioning information. Consequently, each training sample consists of the current vorticity field, the forcing applied during the prediction interval, and the target vorticity at the next snapshot.

\paragraph{Testing.}
For one-step evaluation, we generate 48 conditioning states. Since the forcing over the snapshot interval is observed, the next-snapshot state is deterministic. Each test case therefore consists of a conditioning state, the applied forcing, and the corresponding next-snapshot state. We compare the model ensemble mean with this deterministic target state and its ensemble standard deviation with zero. For the invariant measure and long-term trajectory evaluations, we generate 5,000 trajectories, each consisting of 50 state--forcing pairs.

\begin{table}[h]
  \caption{Physical constants and simulation parameters for the stochastic and deterministic Kolmogorov-flow datasets. The SDE is integrated with solver time step $h$, while snapshots are recorded at intervals of $\Delta t$. The interval $\Delta t$ also defines the model prediction horizon, i.e., the model predicts the snapshot at $t+\Delta t$ from the state at $t$.}
  \label{tab:sim_params}
  \centering
  \begin{tabular}{lcc}
    \toprule
    \textbf{Parameter} & \textbf{Symbol} & \textbf{Value} \\
    \midrule
    Solver time step   & $h$             & $10^{-4}$      \\
    Snapshot interval  & $\Delta t$      & $0.5$          \\
    Solver grid resolution   & $N$             & $256$          \\
    Model grid resolution & $\widehat{N}$   & $64$           \\
    Viscosity          & $\nu$           & $10^{-3}$      \\
    Linear drag coefficient        & $\alpha$        & $0.1$          \\
    Forcing amplitude      & $\epsilon$      & $1.0$          \\
    \bottomrule
  \end{tabular}
  \vskip -0.1in
\end{table}

\section{Baseline Specifications}
\label{app:baselines}

\subsection{Diffusion Models: DDIM, DPM-2, DDPM}
\label{app:diffusion}

DDIM, DPM-2, and DDPM share the same continuous-time $\epsilon_\theta$-prediction model trained with a linear variance-preserving (VP) noise schedule parameterized by $\beta_{\min}$ and $\beta_{\max}$ (Table~\ref{tab:hparams}). The baselines differ only in the sampling procedure:
\begin{itemize}
    \item \textsc{DDIM} corresponds to the $\eta=0$ setting of the DDIM sampler \citep{song2020denoising}, which recovers the first-order exponential integrator for the probability-flow ODE.
    \item \textsc{DPM-2} uses the second-order exponential integrator DPM-Solver-2 \citep{cheng2022dpm}. Unlike the original DPM-Solver, which places integration steps uniformly in log-SNR, we discretize uniformly in time. Appendix~\ref{app:dpm2_discretization} shows that both discretizations perform comparably, with the uniform schedule yielding slightly lower standard deviation errors at low inference budgets.
    \item \textsc{DDPM} corresponds to the $\eta=1$ setting, i.e., ancestral sampling \citep{ho2020denoising}.
\end{itemize}

\subsection{Flow Matching}
\label{app:fm}

The velocity model $v_\theta$ is trained on the conditional OT path with minimum noise level $\sigma_{\min}$ (Table~\ref{tab:hparams}). For sampling, we place Euler steps uniformly in $\tau$.

\subsection{Stochastic Interpolation}
\label{app:si}

We use the training objective of \citet{chen2024probabilistic} with the interpolation schedules $\alpha_\tau = \sigma_\tau = 1-\tau$ and $\beta_\tau = \tau^2$ (Table~\ref{tab:hparams}). Unlike diffusion and flow matching, sampling is initialized from the conditioning state rather than from Gaussian noise, and the model learns the conditional drift as a function of both the conditioning state and the interpolated state. We place Euler--Maruyama steps uniformly in $\tau$, restricting the integration to $\tau \in [0, 0.999]$ to avoid the degenerate noise-free endpoint.

\subsection{Consistency Distillation}
\label{app:cd}

Consistency distillation trains a student model from a pretrained EDM diffusion teacher using the Karras noise schedule $\sigma$ \citep{karras2022elucidating}. The student is initialized with the teacher weights and trained to make consistent predictions, such that neighboring points along the teacher's probability-flow ODE trajectory are mapped to the same clean sample. To this end, we discretize the teacher's noise schedule into $T = 100$ noise levels. For training stability, we use an exponential moving average (EMA) target in the self-consistency objective.

Sampling can either use a single step at the highest noise level or multiple steps at progressively lower noise levels. CD-1 corresponds to single-step sampling using only the highest noise level $\sigma(99)$. For multi-step sampling, we select the noise levels based on the schedule analysis in Appendix~\ref{app:cd_schedules}. In CD-2, the initial prediction at $\sigma(99)$ is refined by a second evaluation at the noise level that is most effective for the specific task: we use $\sigma(97)$ for the stochastic task and $\sigma(50)$ for the deterministic task.

\subsection{Adversarial Diffusion Distillation}
\label{app:add}

ADD combines a frozen teacher, a trainable student generator, and a discriminator. In our implementation, the teacher and generator are flow matching models. The generator learns to match the 10-step teacher prediction in a single step while producing samples the discriminator cannot distinguish from real data.

The discriminator uses the encoder and the bottleneck of the teacher U-Net as a feature extractor. MLP heads are attached to the input layer, each encoder stage, and the bottleneck. Each head applies global average pooling to its feature maps and outputs a scalar logit. The final discriminator score is obtained by summing the hinge losses of all logits. The discriminator is trained to assign higher scores to real samples than to generated samples and is regularized by an R1 gradient penalty (coefficient $\gamma$).

The generator objective combines a distillation term and an adversarial term. The distillation term causes the generator to match the teacher model predictions, while the adversarial term forces the generator to produce realistic samples that receive high discriminator scores. The weight $\lambda$ balances the adversarial and distillation terms, with $\lambda = 1$ corresponding to pure distillation and $\lambda = 0$ to a standard GAN.

The generator and discriminator are trained in parallel, with the generator updated once every $r_{\mathrm{gen}} = 5$ discriminator updates to keep the discriminator ahead and provide a stable training signal. We additionally use a balancing mechanism similar to \citet{berthelot2017began}.

\begin{table}[t]
\caption{Model-specific hyperparameters. The diffusion models share the same continuous-time $\epsilon_\theta$-prediction parameterization and differ only in the sampler used at inference time.}
\label{tab:hparams}
\centering
\small
\renewcommand{\arraystretch}{1.1}
\setlength{\tabcolsep}{4pt}
\begin{minipage}[t]{0.48\columnwidth}
\vspace{0pt}
\centering
\begin{tabular}{@{}lll@{}}
\toprule
\textbf{Parameter} & \textbf{Symbol} & \textbf{Value} \\
\midrule
\multicolumn{3}{@{}l}{\textbf{Diffusion}}\\
Variance schedule
    & $\beta(\tau)$
    & VP linear\\
Schedule limits
    & $[\beta_{\min},\beta_{\max}]$
    & $[0.1,20]$\\
Sampler
    & --
    & \textsc{DPM-2}/\textsc{DDIM}/\textsc{DDPM}\\
\addlinespace
\multicolumn{3}{@{}l}{\textbf{Flow Matching}}\\
Minimum noise level
    & $\sigma_{\min}$
    & $10^{-3}$\\
Integrator
    & --
    & Euler\\
\addlinespace
\multicolumn{3}{@{}l}{\textbf{Stochastic Interpolation}}\\
Conditioning coefficient
    & $\alpha_\tau$
    & $1-\tau$\\
Target coefficient
    & $\beta_\tau$
    & $\tau^2$\\
Noise schedule
    & $\sigma_\tau$
    & $1-\tau$\\
Time range
    & $\tau$
    & $[0,0.999]$\\
Integrator
    & --
    & Euler--Maruyama\\
\bottomrule
\end{tabular}
\end{minipage}
\hfill
\begin{minipage}[t]{0.48\columnwidth}
\vspace{0pt}
\centering
\begin{tabular}{@{}lll@{}}
\toprule
\textbf{Parameter} & \textbf{Symbol} & \textbf{Value} \\
\midrule
\multicolumn{3}{@{}l}{\textbf{Consistency Distillation}}\\
Noise schedule
    & $\sigma$
    & Karras\\
Noise schedule exponent
    & $\rho$
    & $7$\\
Minimum noise level
    & $\sigma_{\min}$
    & $2\times10^{-3}$\\
Maximum noise level
    & $\sigma_{\max}$
    & $80$\\
Noise levels
    & $T$
    & $100$\\
EMA decay
    & --
    & $0.999$\\
\addlinespace
\multicolumn{3}{@{}l}{\textbf{Adversarial Diffusion Distillation}}\\
Teacher steps
    & $T_{\mathrm{teacher}}$
    & $10$\\
Student steps
    & $T_{\mathrm{student}}$
    & $1$\\
Update ratio
    & $r_{\mathrm{gen}}$
    & $5$\\
Distillation weight
    & $\lambda$
    & $0.9$\\
$R_1$ coefficient
    & $\gamma$
    & $1.0$\\
\bottomrule
\end{tabular}
\end{minipage}
\end{table}

\section{Consistency Distillation Sampling Schedules}
\label{app:cd_schedules}

Consistency models support multi-step sampling by re-evaluating the consistency function at progressively lower noise levels. Unlike diffusion- and flow-based samplers, this procedure does not arise from discretizing an underlying SDE or probability-flow ODE. As a consequence, more sampling steps do not necessarily improve sample quality; \citet{chen2025convergence} show that too many steps can even degrade it. This motivates an empirical analysis of sampling schedules for the stochastic and deterministic settings.

\begin{figure}[h]
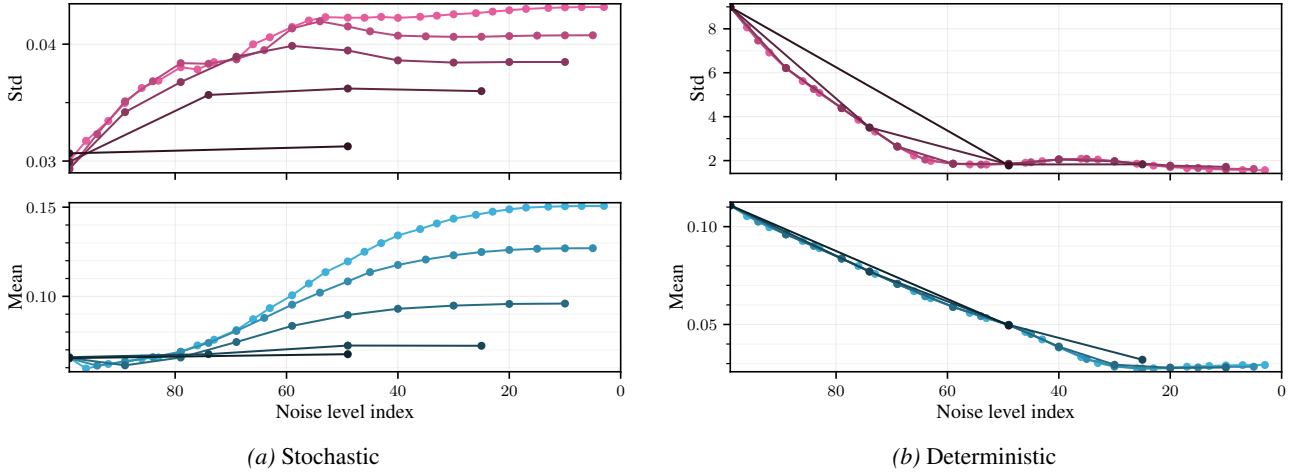

  \centering
  \begin{subfigure}[t]{0.49\textwidth}
    \centering
    \resizebox{\linewidth}{!}{\import{plots/stoc_cd_schedule/}{stoc_cd_schedule.pgf}}
    \caption{Stochastic}
  \end{subfigure}
  \hfill
  \begin{subfigure}[t]{0.49\textwidth}
    \centering
    \resizebox{\linewidth}{!}{\import{plots/det_cd_schedule/}{det_cd_schedule.pgf}}
    \caption{Deterministic}
  \end{subfigure}
  \caption{Evaluation of different Consistency Distillation sampling schedules for stochastic and deterministic Kolmogorov flow. Each curve corresponds to a sampling schedule. The mean and standard deviation errors at each marker are obtained by performing denoising steps at all marked noise levels up to and including that marker.}
  \label{fig:cd_schedules}
\end{figure}

Figure~\ref{fig:cd_schedules} compares different sampling schedules, reporting the mean and standard deviation errors after each denoising step. In the stochastic case, a single step at the highest noise level already performs near optimally. A second step at the high-noise end slightly improves the mean error, while further steps yield no benefit. Schedules that distribute many steps across the full noise range perform worse on both metrics.

In the deterministic case, a single step at the highest noise level is insufficient: steps at lower noise levels are needed to obtain an accurate mean and eliminate residual variability. Here, schedules with evaluations distributed uniformly across the noise range are most effective.

\section{DPM-2 Discretization Schedule}
\label{app:dpm2_discretization}
\begin{figure}[h]
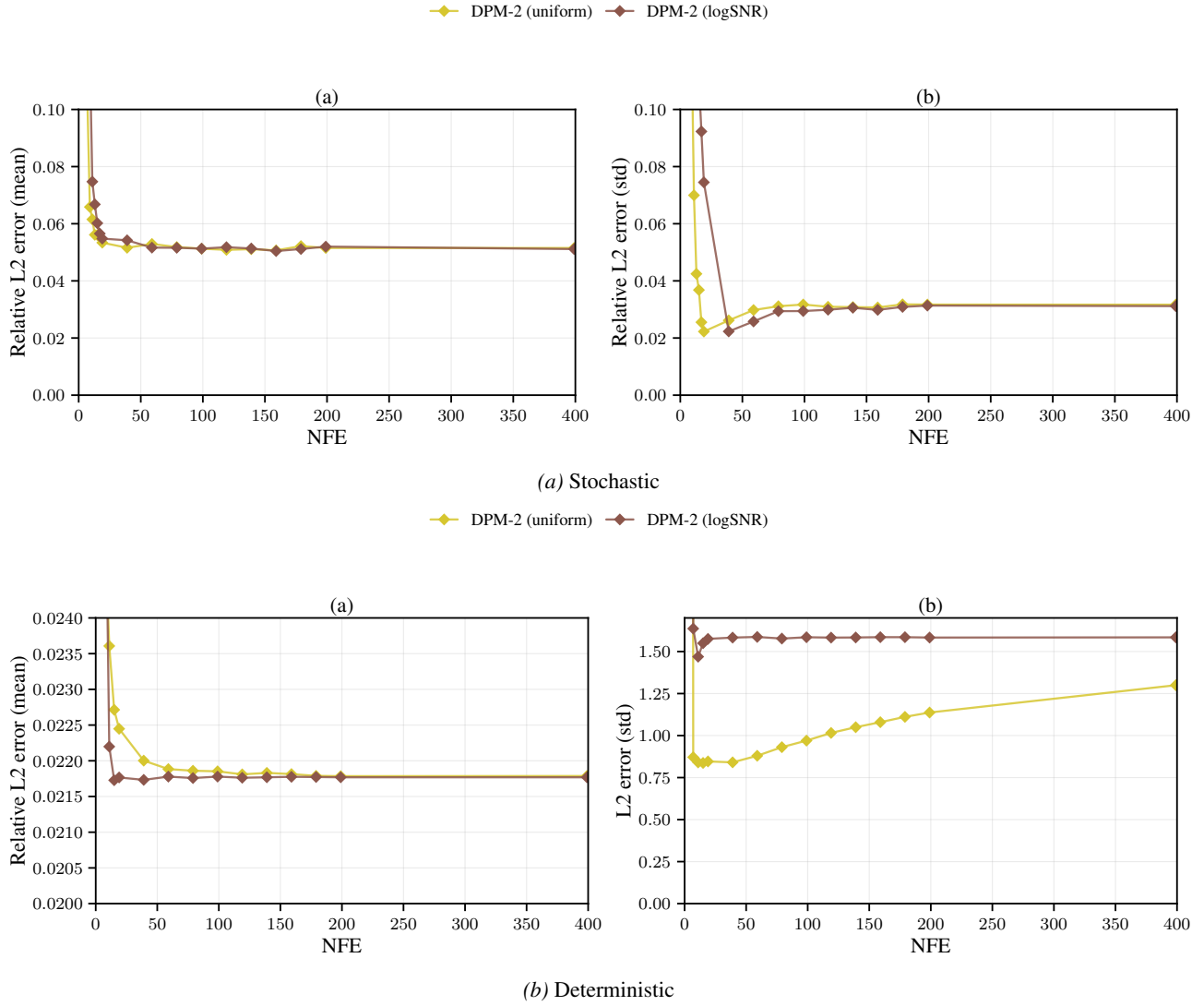

  \centering
  \begin{subfigure}[t]{\textwidth}
    \centering
    \resizebox{\linewidth}{!}{\import{plots/stoc_dpm2_disc/}{stoc_dpm2_disc.pgf}}
    \caption{Stochastic}
    \label{fig:dpm2_disc_stoc}
  \end{subfigure}
  \begin{subfigure}[t]{\textwidth}
    \centering
    \resizebox{\linewidth}{!}{\import{plots/det_dpm2_disc/}{det_dpm2_disc.pgf}}
    \caption{Deterministic}
    \label{fig:dpm2_disc_det}
  \end{subfigure}
  \caption{Uniform vs.\ log-SNR discretization for \textsc{DPM-2} on stochastic and deterministic Kolmogorov flow.}
  \label{fig:dpm2_disc}
\end{figure}

The original DPM-Solver \citep{cheng2022dpm} places integration steps uniformly in log-SNR, which concentrates evaluations in the low-noise regime. Our \textsc{DPM-2} baseline instead discretizes uniformly in time. Figures~\ref{fig:dpm2_disc_stoc} and~\ref{fig:dpm2_disc_det} compare the two discretizations. The mean errors are nearly identical on both tasks. In the stochastic case, the standard deviation errors converge at high budgets, while at low budgets the log-SNR schedule amplifies the overestimation of the standard deviation (cf.\ Figure~\ref{fig:std_norm}, left). In the deterministic case, the uniform schedule has lower residual variability across all inference budgets.

\section{Supplementary Evaluation Figures}
\label{app:figures}

\begin{figure}[H]
  \centering
  \resizebox{0.75\textwidth}{!}{\import{plots/stoc_enstr_spec/}{stoc_enstr_spec.pgf}}
  \caption{Ensemble-averaged radial enstrophy spectra after 50 autoregressive rollout steps on stochastic Kolmogorov flow. Dotted, dashed, and solid lines indicate models using 1, 2, and 400 NFEs, respectively.}
  \label{fig:stoc_enstr_spec}
\end{figure}

\begin{figure}[H]
  \centering
  \resizebox{0.75\textwidth}{!}{\import{plots/det_enstr_spec/}{det_enstr_spec.pgf}}
  \caption{Ensemble-averaged radial enstrophy spectra after 50 autoregressive rollout steps on deterministic Kolmogorov flow. Dotted, dashed, and solid lines indicate models using 1, 2, and 400 NFEs, respectively.}
  \label{fig:det_enstr_spec}
\end{figure}

\begin{figure}[H]
  \centering
  \resizebox{!}{0.95\textheight}{\import{plots/stoc_seed_fields/}{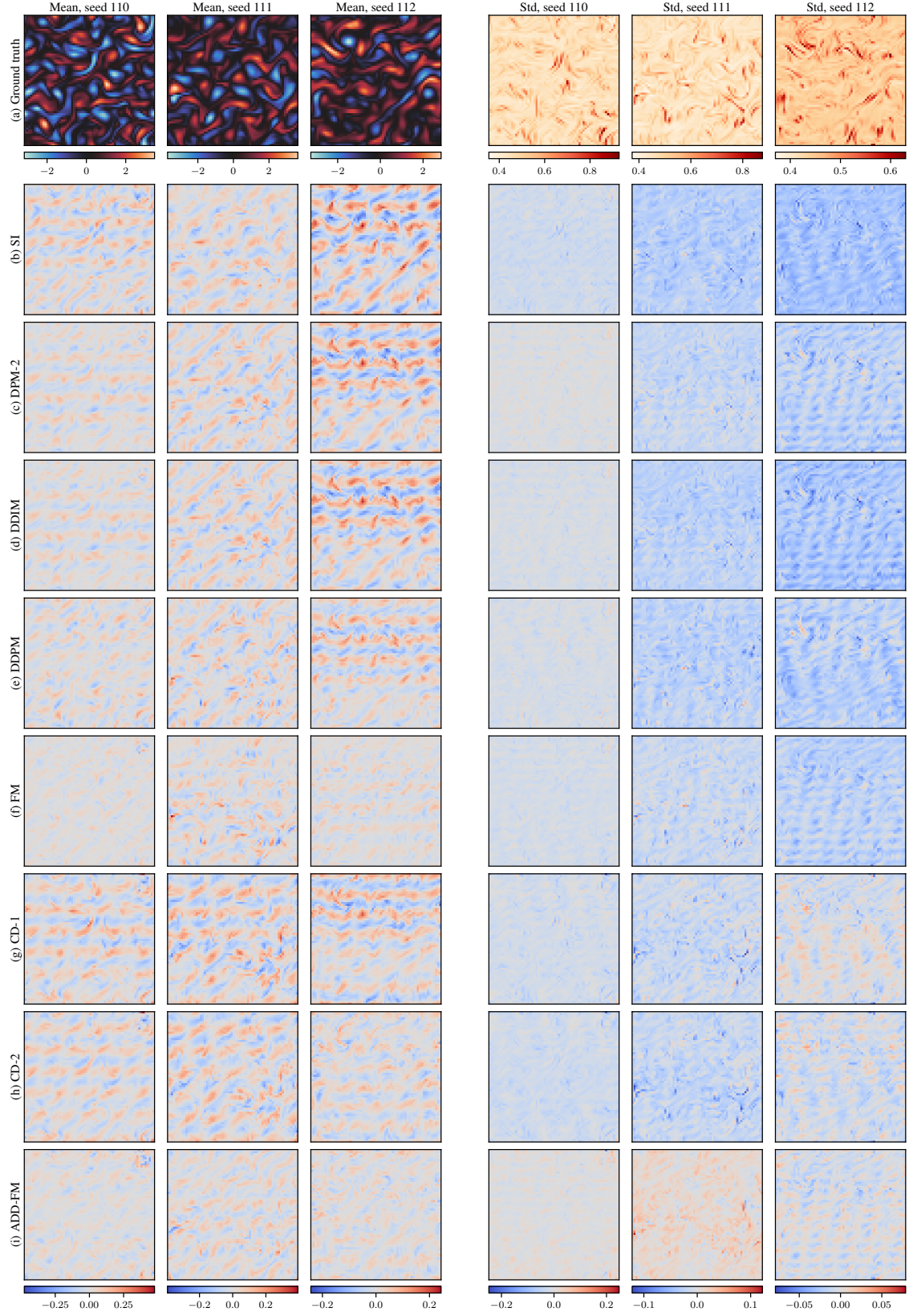}}
  \caption{Conditional means and standard deviations for three stochastic Kolmogorov flow conditions. The top row presents the ground-truth statistics, and the remaining rows show model errors relative to the ground truth. Color scales are shared within each column.}
  \label{fig:stoc_seed_fields}
\end{figure}

\end{document}